\documentclass[letterpaper, 10 pt, conference]{ieeeconf}  

\IEEEoverridecommandlockouts                              

\usepackage{booktabs}       
\usepackage{amsfonts}       
\usepackage{nicefrac}       
\usepackage{microtype}      
\usepackage{float}
\usepackage{mathtools}  
\usepackage[usenames, dvipsnames]{xcolor}
\usepackage{multirow} 
\usepackage{algorithm}
\usepackage{algorithmic}
\usepackage{amsfonts}
\usepackage{mathtools,amssymb}
\usepackage{fixltx2e}
\usepackage{array}
\usepackage{setspace}
\usepackage{caption} 
\usepackage{bbold}
\usepackage{bm}
\usepackage{amsfonts}
\usepackage{hyperref}
\usepackage{url}

\usepackage{expl3}
\ExplSyntaxOn
\newcommand\latinabbrev[1]{
  \peek_meaning:NTF . {
    #1\@}%
  { \peek_catcode:NTF a {
      #1.\@ }%
    {#1.\@}}}
\ExplSyntaxOff

\def\etal{\latinabbrev{et al}}

\def\benchmark{LiDAR-SOT}

\def\tracker{SOTracker}

\title{\LARGE \bf 
Model-free Vehicle Tracking and State Estimation \\ in Point Cloud Sequences} 

\author{Ziqi Pang, Zhichao Li, Naiyan Wang \\ 
        TuSimple \\
      \tt\small \{ziqipang.z, leeisabug, winsty\}@gmail.com}

\begin{document}
\maketitle 
\thispagestyle{empty}
\pagestyle{empty}

\begin{abstract}

Estimating the states of surrounding traffic participants stays at the core of autonomous driving.
In this paper, we study a novel setting of this problem: model-free single-object tracking (SOT),
which takes the object state in the first frame as input, and jointly solves state estimation and tracking in subsequent frames.
The main purpose for this new setting is to break the strong limitation of the popular ``detection and tracking'' scheme in multi-object tracking.
Moreover, we notice that shape completion by overlaying the point clouds, which is a by-product of our proposed task, not only improves the performance of state estimation but also has numerous applications.
As no benchmark for this task is available so far, we construct a new dataset \emph{LiDAR-SOT} and corresponding evaluation protocols based on the Waymo Open dataset~\cite{waymo_open}.
We then propose an optimization-based algorithm called \emph{SOTracker} involving point cloud registration, vehicle shapes, correspondence, and motion priors. 
Our quantitative and qualitative results prove the effectiveness of our SOTracker and reveal the challenging cases for SOT in point clouds, including the sparsity of LiDAR data, abrupt motion variation, etc.
Finally, we also explore how the proposed task and algorithm may benefit other autonomous driving applications, 
including simulating LiDAR scans, generating motion data, and annotating optical flow.
The code and protocols for our benchmark and algorithm are available at \url{https://github.com/TuSimple/LiDAR_SOT/}.
A video demonstration is at \url{https://www.youtube.com/watch?v=BpHixKs91i8}.

\end{abstract}

\section{Introduction}
Autonomous driving calls for accurate state estimation of surrounding objects, including their positions, orientations, etc.
Among all the sensors available for state estimation, LiDAR is unique in providing accurate 3D sensing.
Nevertheless, the complexity of road scenarios and sparsity of point clouds pose great difficulties for state estimation.
Therefore, estimating the states of objects using LiDAR data has long been valuable yet challenging for autonomous driving research.

\begin{figure}
    \centering
    \includegraphics[width=1.0\columnwidth]{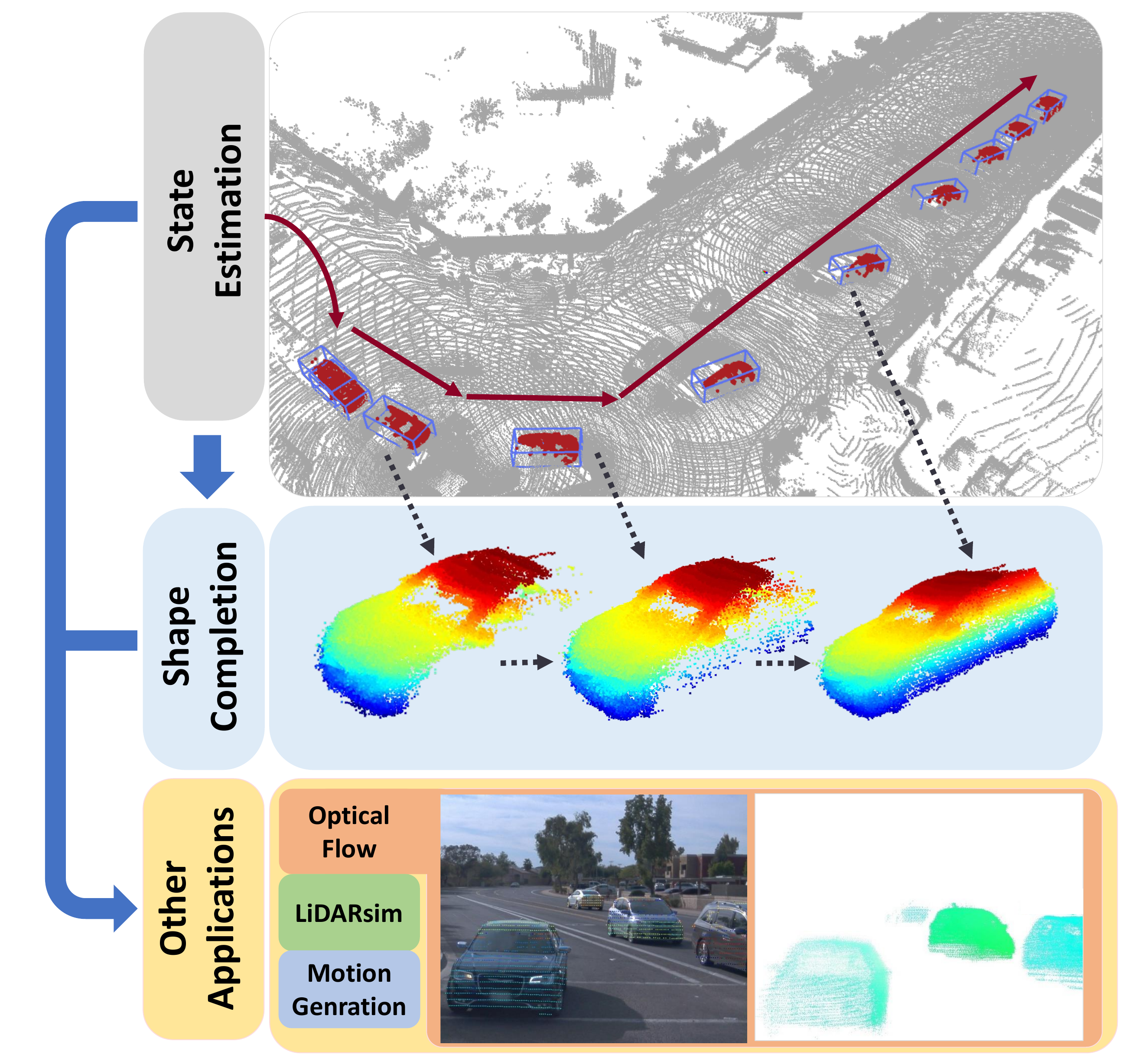}
    \caption{Our vehicle tracking system and its applications. \textbf{Top}. Tracking results  visualized at the interval of 2 seconds (20 frames).
    \textbf{Middle}. Shape completion by overlaying point clouds.
    \textbf{Bottom}. Applications using the estimated states and shapes, such as optical flow annotation.
    More usages are described in Sec.~\ref{sec::downstream}.}
    \label{fig::cover}
    \vspace{-20px}
\end{figure}

The mainstream method for object tracking is to utilize the so-called ``detection and tracking'' paradigm.
It mostly focuses on the association of bounding boxes across frames.
The accuracy of tracking then heavily relies on the performance of detection,
since most tracking algorithms directly accept the bounding boxes from detection with minor modifications.
Current state-of-the-art object detection algorithms are mostly \emph{model-based}:
the algorithms carry out the detection task with their recognition power gained from learning on large scale training sets.
The benchmarks nowadays~\cite{waymo_open}\cite{argoverse}\cite{ KITTI}\cite{nuScenes} also mainly focus on this \emph{model-based} MOT setting.
However, we propose to pay attention to the \emph{model-free} Single-Object Tracking (SOT) setting,
which targets on some overlooked applications and challenges in MOT.

\begin{figure}
    \centering
    \includegraphics[width=1.0\columnwidth]{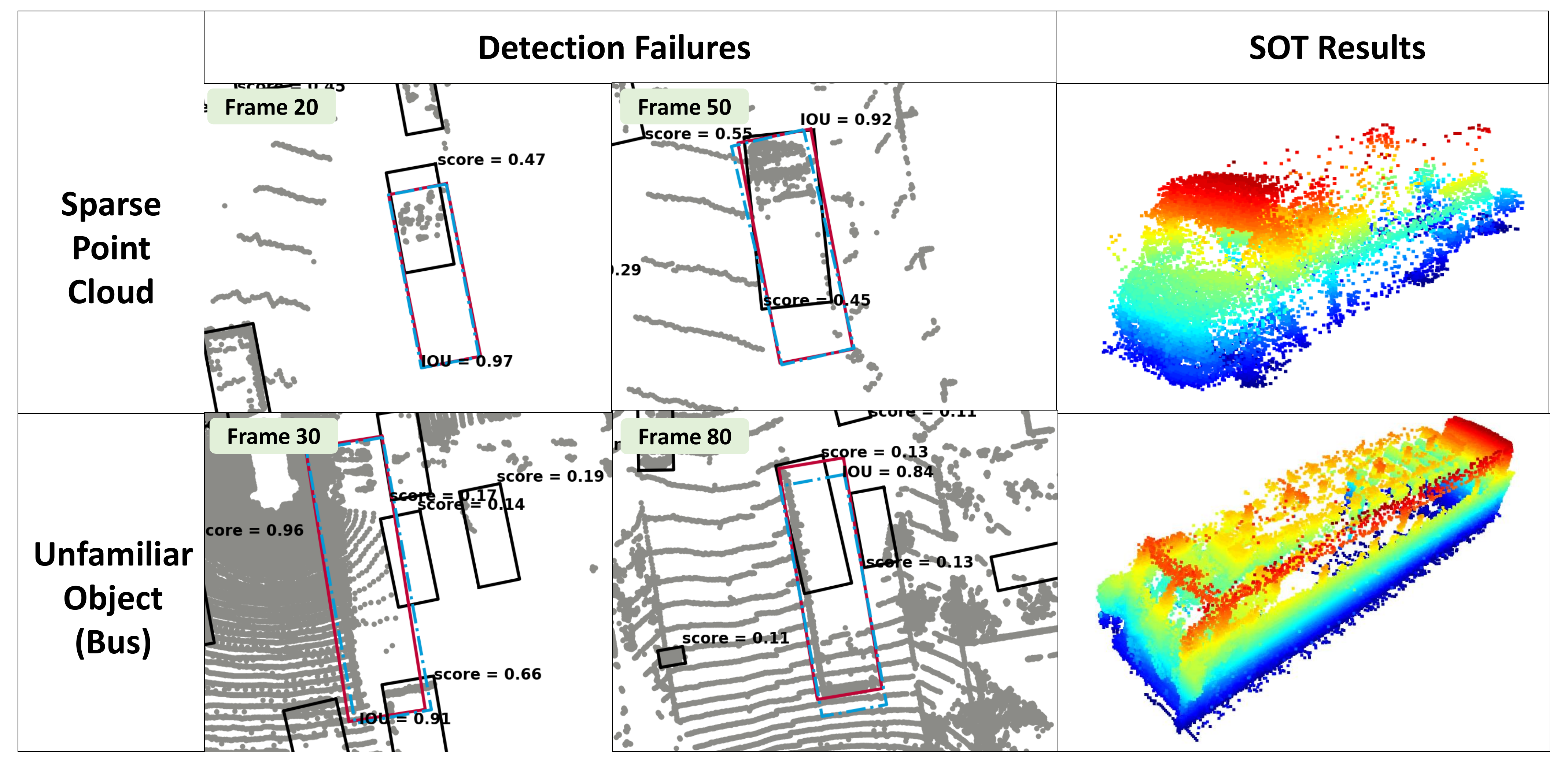}
    \caption{\textbf{Left:} The detection algorithm (PointPillars~\cite{pointpillars}) fails on certain cases,
        including false negatives and wrong size. 
        But our tracker still has high quality after a period of tracking, with only requiring an initial bounding box as input.
    	We sample two frames from a continuous period of failed detection for demonstration.
    \emph{Gray}: Point Cloud; \emph{Red}: The Ground Truth Box; \emph{Dashed Blue}: SOT Box; \emph{Black}: Detection Boxes.
    \textbf{Right:} We visualize the quality of SOT with the aggregated point clouds, following~\cite{Held-Combining}\cite{IROS-Continuous-time}.
    Our algorithm can provide accurate tracking and shape aggregation results.
    }
    \label{fig::large_car}
    \vspace{-20px}
\end{figure}

First, model-based detection is prone to fail on sparse point clouds and unfamiliar objects (see Fig.~\ref{fig::large_car}). By being model-free, the algorithms can generalize to such cases, since it is not assumed that similar objects have been met before. 
The task of \emph{model-free} SOT is then useful for both online and offline applications from two aspects:
1) For offline settings, model-free SOT can augment the training set for model-based methods by providing cheap annotations or simulated data~\cite{LiDARsim}.
2) For online settings, SOT serves as a back-up system for model-based methods on their failures (see Fig.~\ref{fig::large_car}). This feature is mandatory for a high-level autonomous system which requires no single failure point. 

Second, model-free SOT algorithms rely on richer information for tracking, such as raw point clouds, compared to MOT methods that mostly use bounding boxes only.
Therefore, the SOT algorithms specialize at providing more subtle state information with even centimeter level accuracy.
This property is important for many applications in autonomous driving, such as generating first order motion information. It is discussed in Sec.~\ref{subsec::motion_generation}.

For a target object, the input to SOT algorithms is its initial bounding box in the first frame, which may come from either human annotations or highly confident detections.
The algorithms then estimate the object's states in every subsequent frame, as is in the top of Fig.~\ref{fig::cover}.
During solving state estimation, we notice the value of its by-product: aggregating point clouds along the tracklets completes denser shapes of objects (shown in the middle of Fig.~\ref{fig::cover}).
In our experiments, leveraging these shapes improves the performance of tracking. This utility of shape aggregation also benefits a series of other applications in autonomous driving.
For example, with the generated shapes and motion information,
we can create vehicle bank for creating virtual LiDAR scenes~\cite{LiDARsim}, optical flow annotations~\cite{KITTI-Flow} (our result visualized in Fig.~\ref{fig::cover}, bottom row), etc.
These tasks so far mostly require manual labeling or artificial CAD models,
so automating the process of directly using raw point cloud data could greatly improve their efficiency and scalability.
The details of such downstream applications are discussed in Sec.~\ref{sec::downstream}.

To summarize, our contributions are as follows:
\vspace{-5px}
\begin{itemize}
    \item We propose a novel LiDAR-based object tracking task\footnote{We focus on vehicle so far, and are working on extending to other types of objects.}.
    To facilitate further study, we create a new benchmark called \benchmark\ based on the WOD dataset~\cite{waymo_open}.
    \item We propose a model-free SOT algorithm called \tracker. It involves optimizing for point cloud registration, object shapes and motions.
    \item We evaluate our \tracker\ on \benchmark\ and discuss the potential usages in autonomous driving, such as motion data generation, LiDAR scan simulation~\cite{LiDARsim}, and optical flow annotation~\cite{KITTI-Flow}.
\end{itemize}
\section{Related Work}
\label{sec::related-work}

\subsection{Datasets and Benchmarks}
\label{subsec::related-work-benchmarks}
\vspace{-3px}

Many existing autonomous driving benchmarks, including Waymo Open Dataset (WOD)~\cite{waymo_open}, Argoverse~\cite{argoverse}, KITTI~\cite{KITTI}, and nuScenes~\cite{nuScenes}, have rich point clouds and tracklets for object tracking.
However, they mainly adopt the MOT setting while our task demands the SOT setting, and not all the tracklets for MOT evaluation suits the requirements for SOT.
So we design a test bed specially for SOT evaluation.
Among the mentioned benchmarks, WOD~\cite{waymo_open} is superior in scale and comprehensiveness, so we build the benchmark \benchmark\ based on the data of WOD.

To evaluate the SOT algorithms, we inherit the idea from VOT~\cite{VOT} for the tracking performance on 2D images, concerning both the accuracy and robustness.
As for shape completion, we consider it for evaluation because of its usages on downstream applications.
Specifically, we refer to ShapeNet~\cite{shapenet} and use Chamfer-Distance(CD) as the metric.

\vspace{-3px}
\subsection{State Estimation and Tracking in Point Cloud}
\label{subsec::related-work-estimation}
\vspace{-3px}

State estimation in raw point clouds is challenging because it involves associating the LiDAR points with the target object (a.k.a. tracking) and then accurately estimate the state.

Many methods adopt a model-based approach to discover the object in LiDAR points.
\cite{darpa-tracking}\cite{darms2008classification}\cite{tracking-and-detection-icra09}\cite{Generative-object-detection}
fit the vehicles with boxes and~\cite{TanzmeisterTWB14}\cite{ DanescuON11} utilizes occupancy grid for detection.
In the era of deep learning, the learned neural networks replace the manually designed models.
P2B~\cite{P2B} generates and verifies bounding boxes guided by a learned PointNet++~\cite{pointnet++}, 
Zou \textit{\etal}~\cite{zou2020f} manages to fuse RGB information using frustums, 
Zarzar \textit{\etal}~\cite{PointRGCN} refines the detection results by association,
while~\cite{PIXOR}\cite{fast-and-furious} apply convolutional neural networks(CNN) on the Bird-Eye-View images of LiDAR,
and Giancola \textit{\etal}~\cite{LeveragingShapeCompletion} learns to complete the partial point cloud, and then searches for it.

These model-based approaches heavily rely on detection or shape completion in every single frame.
So point cloud sparsity and unfamiliar objects could easily lead to failures.
However, our model-free setting seeks to tackle this since it does not need to fit a model and can consider multi-frame information from LiDAR sequences.

Different from the aforementioned works involving searching for the target object in raw LiDAR data, some assume the point cloud could be well-segmented or the segmentation information is available.
Hence their focus is state estimation in relatively clean point cloud data.
Some representative methods are as follows,
Held \textit{\etal}~\cite{Held-Combining} minimizes an energy function on the inconsistency between observed point cloud and object states.
Gro{\ss} \textit{\etal}~\cite{alignnet-3d} registers the motion between point cloud snapshots.
Goforth \textit{\etal}~\cite{joint-lucey} learns to complete the shapes of vehicles and jointly decode the states.
The problem of these methods lies in the assumption of well-segmented point cloud, which is rare in practice.
Therefore, by only providing the first frame annotation, our setting enforces the algorithms to face the challenge of solving tracking and state estimation simultaneously.

Many model-free works~\cite{moosmann2013joint}\cite{ wang2015model}\cite{dewan2016motion}\cite{pomerleau2014long}\cite{azim2012detection}\cite{tipaldi2009motion}\cite{van2010integrated} 
mainly focus on the problem of MOT.
They first operate object detection leveraging either the motion information or assuming that LiDAR points draw the boundaries of objects. Then they associate the detected bounding boxes according to the estimated motion information or motion models.  
Therefore, these methods differ from our objectives for using SOT. 

The only work we found falls into our setting is Ushani \textit{\etal}~\cite{IROS-Continuous-time}.
Dating back to that period, there is no large-scale dataset fits the evaluation needs,
thus~\cite{IROS-Continuous-time} only tests on a private dataset, which makes the comparisons infeasible.
This is just the reason to propose our new LiDAR-SOT benchmark, which aims to promote the studies in this area.
\vspace{-3px}
\subsection{Point Cloud Completion}
\label{subsec::related-work-completion}
\vspace{-3px}

Following the paradigm of PCN~\cite{PCN}, researchers can complete point clouds with neural networks trained from artificial CAD models in graphics datasets like ShapeNet~\cite{shapenet}.
~\cite{joint-lucey}\cite{LeveragingShapeCompletion} adopt the same idea for vehicles.
However, these methods could hardly scale up as CAD models are expensive and limited to some categories.
Recently, Gu \textit{\etal}~\cite{weakly-completion} discards the need for CAD models by using multi-view sensors.
But multi-view sensor information are also not universal for real-world scenarios.

Considering the mentioned drawbacks of supervised methods, completing point cloud by tracking interests us most.
\cite{DaraeiVM17}\cite{christie20163d}\cite{JiangCPD17} jointly use LiDAR and camera for this task.
~\cite{Held-Combining}\cite{IROS-Continuous-time} are closest to our objective in using LiDAR only:
Held \textit{\etal}~\cite{Held-Combining} aggregates the observed point clouds, while Ushani \textit{\etal}~\cite{IROS-Continuous-time} directly optimizes a point cloud model.
\vspace{-3px}
\section{LiDAR-SOT Benchmark}
\label{sec::benchmark}
\vspace{-3px}
\subsection{Problem Formulation}
\label{subsec::prob_formulation}
\vspace{-3px}

Suppose the dataset has $N$ tracklets $\mathcal{T}^{1:N}=\{\mathcal{T}^{1}, \mathcal{T}^{2}, ...\mathcal{T}^{N}\}$ with lengths $L^{1:N}=\{L^{1}, L^{2}, ..., L^{N}\}$. We refer to the $i$-th frame in the $j$-th sequence $\mathcal{T}^j$ as $\mathcal{T}^{j}_{i}$,
which contains the point cloud $P^{j}_{i}$ and a bounding box $\widetilde{B}^{j}_{i}$. The box $\widetilde{B}^{j}_{i} \triangleq \{\widetilde{C}^j_i, \widetilde{S}^{j}_{i}\}$ describes the size and state of an object. In our benchmark, $\widetilde{C}_i^j$ are the length, width and height of a 3D cuboid, and the state $\widetilde{S}_{i}^{j}\triangleq[\widetilde{x}_i^j, \widetilde{y}_i^j, \widetilde{z}_i^j, \widetilde{\theta}_i^j]$ represents the 3D coordinates of the object's center and the heading. We only consider 4DOF instead of 6DOF because roll and pitch deviations are usually aligned to the road for autonomous driving. For a tracklet $\mathcal{T}^j$, it also contains the shape ground-truth of target object $\widetilde{M}^j$.

For an arbitrary tracklet $\mathcal{T}^j$, the algorithms are fed with point cloud $\{P_{1:L^j}^j\}$ and the ground-truth bounding box $\widetilde{B}_1^j$ in the first frame. The outputs are the estimated states $\{S_{2:L^{j}}^{j}\}$ and completed object shape $M^j$. Note that we assume the size of an object remains unchanged, thus it keeps the value as $\widetilde{C}_{1}^{j}$ in subsequent frames. This assumption is generally held for rigid objects such as vehicles.

The algorithms can run under either online or offline setting.
The online setting restricts the algorithms to access point cloud data $\{P_{1:L^{j}}^L\}$ by order, and output the estimation results before getting the next frame input. The offline setting, in comparison, poses no constraints. In this paper, we concentrate on the online setting because it fits both on-the-road autonomous driving and offline applications.

\vspace{-3px}
\subsection{Discussion with Closely Related Tasks}
\vspace{-3px}

In this section, we highlight the differences between our proposed task and several relevant tasks.


\noindent
\textit{Multiple-object Tracking:} 
As mentioned above, MOT relies on detections and mainly focuses on associating them. 
Moreover, MOT usually does not modify or generate new bounding boxes, while our SOT task creates bounding boxes according to the point cloud and motion information.

\noindent
\textit{Shape Completion:} 
Generally speaking, shape completion complements the partial point cloud in a single frame under the supervised setting. 
It needs paired training data to learn generative models, while ours could utilize temporal LiDAR data to complete the shapes in an unsupervised manner.

\noindent
\textit{Point Cloud Registration:}
Registration focuses on matching the point clouds between two frames, while more previous frames are accessible in our task. 
Consequently, rich priors could be incorporated to improve the performance, as demonstrated later in the experiments, the performance of our state estimation is significantly higher than using ICP only.
\vspace{-3px}
\subsection{Dataset Construction}
\vspace{-3px}

To facilitate the study on the aforementioned task, we propose a new dataset called \benchmark. It is largely based on the most comprehensive and largest point cloud dataset: Waymo Open Dataset (WOD)~\cite{waymo_open}. Considering that the vehicles are the most abundant rigid objects for autonomous driving, we only concern the type of vehicles for now. The WOD dataset contains full annotations of 3D bounding boxes and their associations across frames for the evaluation of MOT, thus, enabling our evaluation for state estimation in point cloud sequences. As for the evaluation for shape completion, since WOD contains no ground truth, we create pseudo-ground-truth shapes (PGTs) for each tracklet by selecting and aggregating the LiDAR points from every frame guided by the ground-truth bounding boxes.

Our algorithm in the paper requires no additional training sets. However, we still include a training set for \benchmark\ by inheriting all the sequences from WOD training set and leaving users with freedom of training/validation split. As for the test set, we select the suitable tracklets from WOD validation set to ensure meaningful evaluation. In total, 1121 vehicle tracklets are included based on the following criteria:

\noindent
\textbf{Tracklet length.} Short sequences may result in unstable evaluation and they are also not challenging enough for SOT tasks.
We therefore filter out the tracklets less than 100 frames.

\noindent
\textbf{Initial number of points.} 
The numbers of LiDAR points in the beginning frames are crucial for the meaningful initialization, where VOT~\cite{VOT} sets up a warm-up period of 10 frames. Enough number of LiDAR points is also necessary for human annotators to provide a initial bounding box with high quality for offline applications.
We therefore take both aspects into consideration, and only keep the tracklets whose initial 10 frames all have more than 20 points. 

\noindent
\textbf{Vehicle mobility.}
We exclude the tracklets in which the vehicle stays static all the time, as these cases have trivial solutions and bring negative biases to the evaluation.
Note that we still keep the tracklets where the vehicles start or end with static states.

\begin{figure}
    \centering
    \includegraphics[width=0.9\columnwidth]{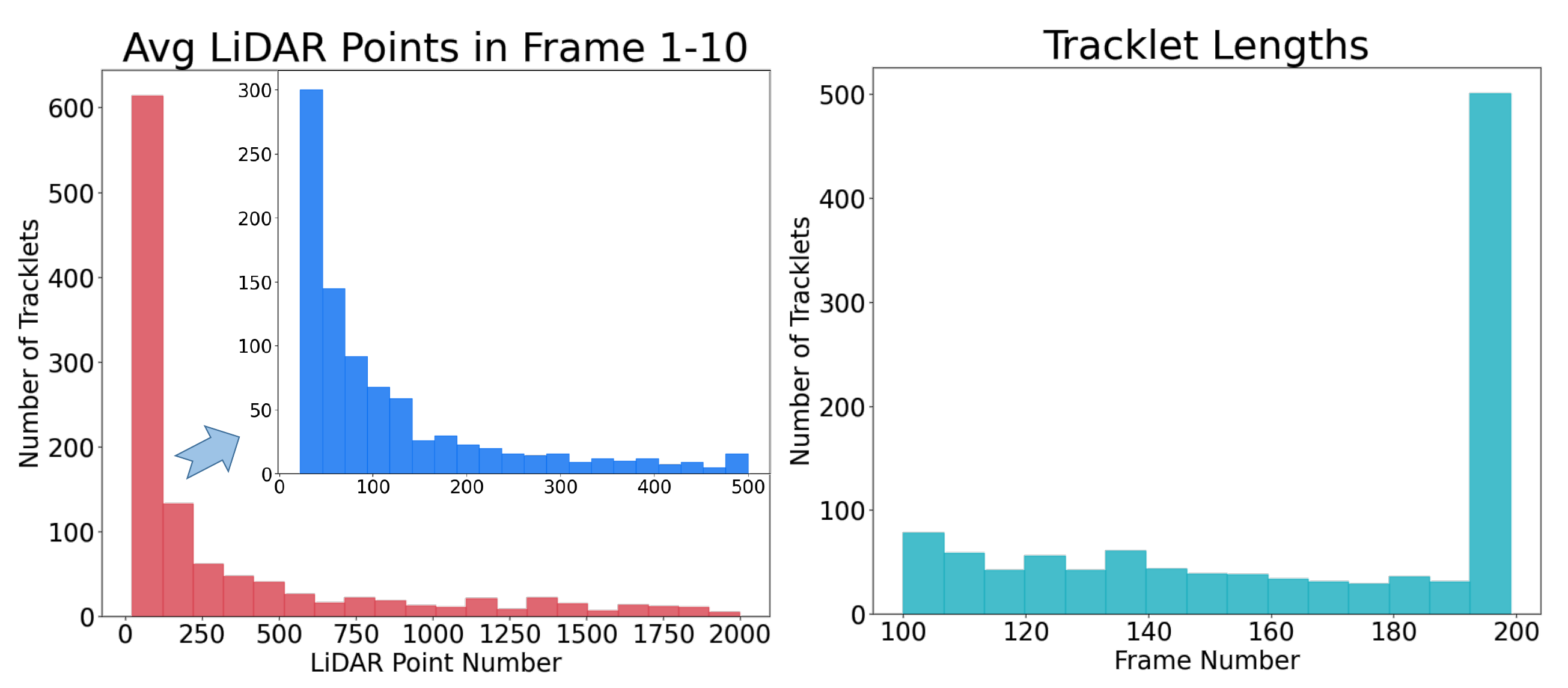}
    \caption{Statistics of \benchmark. 
        \textbf{Left}. Average per-frame LiDAR point number in the first 10 frames.
        \textbf{Right}. Distribution of tracklet lengths. }
    \vspace{-20px}
    \label{fig::benchmark}
\end{figure}

As visualized in the left of Fig.~\ref{fig::benchmark}, although we guarantee the number of LiDAR points in the beginning, the long-tail distribution still indicates the challenges in point cloud sparsity. Moreover, the right of Fig.~\ref{fig::benchmark} illustrates the difficulty of \benchmark\ in long tracklets (note that the original WOD~\cite{waymo_open} has a maximum of 200 frames).

For future analyses, we follow the paradigm of KITTI~\cite{KITTI} and divide the test set into easy, medium and hard subsets.
Intending on the difficulty of point cloud sparsity, we compute the average point number of each tracklet's first 10 frames, then equally split the tracklets into three sets according to the average point number.
The specific thresholds are 38.2 and 808.3 points per frame.

\subsection{Evaluation Metrics}
\label{subsec::metrics}

Regarding the output on tracklet $\mathcal{T}^{j}$: states $\{S_{2:L^{j}}^{j}\}$ and shape $M^j$,
we use multiple metrics for thorough evaluation.
Our focus is on the quality of estimated states in SOT and shape completion.

\noindent
\textbf{State Estimation:}
For state estimation, we are interested in how long the algorithm could track objects and its quality. 
Consequently, inspired by Visual Object Tracking (VOT)~\cite{VOT} work, we propose two metrics for single pass evaluation.

\noindent\textbf{Accuracy:}
We denote $\mathtt{IOU}(B^{j}_{i}, \widetilde{B}^{j}_{i})$ as the $\mathtt{IOU}$ of the estimated and ground-truth bounding boxes.
We then propose a metric reflecting the state estimation accuracy by averaging the $\mathtt{IOU}$ on each frame as: 
\begin{equation}
    \begin{small}
        \begin{aligned}
            \mathtt{Acc}=\frac{1}{\sum_{k=1}^{N}{L^{k}}}\sum_{k=1}^{N}{\sum_{j=1}^{L^{k}}{\mathtt{IOU}(B^{k}_{j}, \widetilde{B}^{k}_{j})}}            
        \end{aligned}
    \end{small}
	\label{eq::metric_overall}
\end{equation}

\noindent\textbf{Robustness:}
Given a pre-defined failure threshold $t$, we let the tracking length $l^{j}(t)$ be the first frame number before $\mathtt{IOU}$ drops lower than $t$ in tracklet $\mathcal{T}^j$.
This can be served as measurement for robustness.
However, as a single threshold $t$ may not be comprehensive, we compute the area under $t$ - $\mathtt{Rob}(t)$ curve by Eq.~\ref{eq::metric_rob}. 
In implementation, we approximate the integral by sampling $t$ from 0 to 1 with intervals of 0.05.
\begin{equation}
    \begin{small}
        \begin{aligned}
    \mathtt{Rob}(t) = \frac{\sum_{k=1}^{N}{l^{k}(t)}}{\sum_{k=1}^{N}{L^{k}}}, \quad \mathtt{Rob} = \int_0^1 \mathtt{Rob}(t)\mathrm{d}t     
        \end{aligned}
    \end{small}
    \label{eq::metric_rob}
\end{equation}

\noindent
\textbf{Shape Completion}
As the PGTs comes from overlaying all the point cloud from tracklets, their densities are non-uniform.
To avoid unstable evaluation, we downsample both the PGTs $\{\widetilde{M}^{1:N}\}$ and our completion $\{M^{1:N}\}$ to the resolution of 5cm during evaluation.
For shape $M^{i}$, we use Chamfer Distance (CD) to measure its distance to $\widetilde{M}^{i}$ as Eq.~\ref{eq::metric_cd}. 
Then the $\mathtt{Shape}$ metric is the average of $\mathtt{CD}$ on all tracklets.
\begin{equation}
	\begin{small}
    \begin{aligned}
        \mathtt{CD}^i = & \frac{1}{|M^i|}\sum_{a\in M^i}\min_{b\in \widetilde{M}^i}||a-b||_2 + 
        \frac{1}{|\widetilde{M}^{i}|}\sum_{b\in \widetilde{M}^i}\min_{a\in M^i}||a-b||_2 \label{eq::metric_cd} \\
    \end{aligned}    
    \end{small}
\end{equation}
\vspace{-8px}
\vspace{-15px}
\section{SOTracker}
\label{sec::method}
In this section, we elaborate our proposed \tracker. 
The SOT task is essentially a cross frame registration task, but a series of priors could be explored for enhancement. 
Briefly speaking, our method estimates the relative motions between frames using registration and various motion priors, and aggregate them. 
Along the process of tracking, we also combine the shapes on the fly for further improvement.

\vspace{-3px}
\subsection{Pipeline}
\label{subsec::pipeline}
\vspace{-3px}

We take an arbitrary tracklet $\mathcal{T}$ for example to explain the algorithm.
The notation $\Delta S_{k} = S_k - S_{k-1}$ is the motion between $S_{k-1}$ and $S_k$. 
Its entries $\Delta S_{k}=[\Delta x_k, \Delta y_k, \Delta z_k, \Delta \theta_k]$ are the 3D translation and rotation of heading.
In the following sections, we use $O_{k}$ to denote the estimated LiDAR points on the surface of the object in frame $\mathcal{T}_k$, in comparison with the raw point cloud $P_k$. 

The pipeline of \tracker\ is in Algo.~\ref{algo::track}, where
it successively estimates the state in each frame with function $\mathtt{FrameEstimate}$ in Algo.~\ref{algo::iter}.
In frame $\mathcal{T}_k$, \tracker\ first solves $S_k$ with $\mathtt{FrameEstimate}$; 
then forms shape $M_k$ by combining previous shape $M_{k-1}$ and object point cloud $O_{k}$ together;
it eventually updates the motion information $\Delta S_{prior}$, which is used for initializing the next frame's state estimation. 
In $\mathtt{FrameEstimate}$, the function initializes the state $\widehat{S}_k$ from adding the $\Delta S_{prior}$ onto previous state $S_{k-1}$.
Then it iteratively selects the relevant LiDAR points $\widehat{O}_{k}$ and optimizes the state $S_{k}$ towards a loss function $\mathcal{L}$. 

Some other details are clarified as follows:

\subsubsection{}We implement $\Delta S_{prior}$ as the moving average of motions and use $\alpha=0.5$ for the update in Algo.~\ref{algo::track}.

\subsubsection{}$\mathtt{PointInBox}(P, S, C, \gamma)$ selects LiDAR points from $P$ that locates in the cuboid with center $S$ and size $\gamma C$.
We set different $\gamma$ values for different usages.
$\gamma_{in}=1.5$ is for the selection inside an estimation loop, while $\gamma_{aft}=1.1$ is for updating shapes.

\subsubsection{}The shape update happens in $\mathtt{OverlayShape}$, where we initialize it using all the points in $\widetilde{S}_1$.
As taking LiDAR points from every frame brings heavy burden for computation, 
we only combine new point clouds at key frames (that is every 5 frames in our implementation).

\subsubsection{}The second frame is special for $\mathtt{FrameEstimate}$ as previous motion information $\Delta S_{prior}$ is not available.
During estimating $S_2$, we let $\Delta S_{prior}=0$ and use a larger search region for $\mathtt{PointInBox}$ by setting $\gamma_{in}=3.0$.
After this, we initialize the motion prior by $\Delta S_{prior}\xleftarrow{}\Delta S_2$.

\begin{algorithm}[!]
    \renewcommand{\algorithmicrequire}{\textbf{Input:}}
	\renewcommand{\algorithmicensure}{\textbf{Output:}}
    \caption{Algorithm for tracklet $\mathcal{T}$}
    \label{algo::track}
    \begin{algorithmic}[1]
        \REQUIRE Point cloud: $P_{1:L}$, bounding box $\widetilde{B}_1$, 
        Size Scaling: $\gamma$,
        Motion Prior Update Factor: $\alpha$ 
        \ENSURE vehicle states: $S_{2:L}$, shape $M_{L}$
        \STATE Frame number $k \xleftarrow{} 2$
        \STATE $\Delta S_{prior} \xleftarrow{} 0$
        \STATE $O_1 \xleftarrow{} \mathtt{PointInBox}(P_1, \widetilde{S}_1, \widetilde{C}_1, 1.0)$
        \STATE $M_1 \xleftarrow{} O_1$
        \FOR{$k$ $<=$ $L$}
        \STATE $S_{k}\xleftarrow{}\mathtt{FrameEstimate}(P_k, O_{k-1}, \Delta S_{prior}, S_{k-1},$\\
        \qquad\qquad\qquad\qquad$\widetilde{C}_1, M_1, M_{k-1}, \gamma_{in})$
        \STATE $O_{k} \xleftarrow{} \mathtt{PointInBox}(P_k, S_K, \widetilde{C}_1, \gamma_{aft})$
        \STATE $M_{k} \xleftarrow{}\mathtt{OverlayShape}(O_{k}, S_{k}, M_{k}, k)$
        \STATE $\Delta S_{prior} \xleftarrow{} \alpha\Delta S_{prior} + (1-\alpha)\Delta S_{k}$
        \STATE $k \xleftarrow{} k + 1$
        \ENDFOR
    \end{algorithmic}
\end{algorithm}
\begin{algorithm}[!]
    \renewcommand{\algorithmicrequire}{\textbf{Input:}}
	\renewcommand{\algorithmicensure}{\textbf{Output:}}
    \caption{$\mathtt{FrameEstimate}$ on frame $\mathcal{T}_k$}
    \label{algo::iter}
    \begin{algorithmic}[1]
        \REQUIRE Point Cloud $P_{k}$ and $O_{k-1}$, Motion Prior: $\Delta S_{prior}$, State: $S_{k-1}$, Size $\widetilde{C}_1$, 
        Shape $M_1, M_{k-1}$, parameters $\gamma_{in}$
        \ENSURE State Estimation $S_k$
        \STATE $\widehat{S}_{k} \xleftarrow{} S_{k-1} + \Delta S_{prior}$
        \FOR{Iteration Number $<$ MaxIter}
        \STATE $\widehat{O}_{k} \xleftarrow{} \mathtt{PointInBox}(P_{k}, \widehat{S}_k, \widetilde{C}_{1}, \gamma_{in})$
        \STATE $S_{k}^*\xleftarrow{} \mathop{\arg\min}\limits_{S_{k}} \mathcal{L}(S_{k}, \widehat{O}_k, O_{k-1}, S_{k-1}, M_{k-1}$)
        \STATE $\widehat{S}_{k} \xleftarrow{} S_{k}^*$
        \ENDFOR
        \STATE $S_{k} \xleftarrow{} \widehat{S}_{k}$, 
    \end{algorithmic}
\end{algorithm}

\vspace{-3px}
\subsection{Objective Function} 
\vspace{-3px}
\label{subsec::factor}

The objective function is in Eq.~\ref{eq::loss}. It is the weighted sum of the four terms: ICP Term, Shape Term, Motion Consistency Term, and Motion Prior Term.
We set their weights by $[1, 1, 0.1, 0.1]$.
Eventually, we use the BFGS solver from scipy~\cite{virtanen2020scipy} to optimize $\mathcal{L}$.
\begin{equation}
    \label{eq::loss}
    \begin{small}
        \begin{aligned}
            \mathcal{L} = w_{I}\mathcal{L}_{ICP} + w_{S}\mathcal{L}_{Shape} + w_{MC}\mathcal{L}_{MC} + w_{MP}\mathcal{L}_{MP}            
        \end{aligned}
    \end{small}
\end{equation}

In the sequel, we use the notation $\Delta S \otimes (P, S)$ to denote apply transformation $\Delta S=[\Delta x, \Delta y, \Delta z, \Delta \theta]$ to the point cloud $P$, relative to the coordinate system centered at $S=[x, y, z ,\theta]$.
$\Delta S \otimes (P, S)$ equals to the first 3 rows of Eq.~\ref{eq::motion_notation}.

\begin{equation}
\begin{small}
	\begin{aligned}
	\begin{bmatrix}
		\cos\Delta\theta & -\sin\Delta\theta & 0 & x+\Delta x-x\cos\Delta\theta+y\sin\Delta\theta\\
		\sin\Delta\theta &  \cos\Delta\theta & 0 & y+\Delta y-x\sin\Delta\theta-y\cos\Delta\theta \\
		0                & 0                 & 1 & \Delta z \\
		0                & 0                 & 0 & 1
	\end{bmatrix}
	\begin{bmatrix}
        P \\
        1
    \end{bmatrix}
	\end{aligned}
\end{small}
\label{eq::motion_notation}
\end{equation}

We explain each term in the following sections. 

\subsubsection{ICP Term}
\label{subsubsec::icp}

Registering the point cloud $\widehat{O}_k$ and $O_{k-1}$ indicates the motion between these two frames.
Therefore, we propose a registration term $\mathcal{L}_{ICP}$ as in Eq.~\ref{eq::icp}. Note that $p_{S_{k-1}}$ indicates applying transformation to $p$ relative to $S_{k-1}$, as Eq.~\ref{eq::motion_notation}.
The details for finding associated point pairs $\mathop{\mathcal{A}}_{I}^{O_{k - 1}, \widehat{O}_k}$ are in Sec.~\ref{subsubsec::lidar_pa}.

\begin{equation}
    \begin{footnotesize}
        \begin{aligned}
    \mathcal{L}_{ICP}=\frac{1}{|\mathop{\mathcal{A}}_{I}^{O_{k - 1}, \widehat{O}_k}|}\sum_{(p, q) \in \mathop{\mathcal{A}}_{I}^{O_{k - 1}, \widehat{O}_k}}||\Delta S_k \otimes (p, S_{k-1})-q||_2^2
        \end{aligned}
    \end{footnotesize}
    \label{eq::icp}
\end{equation}

\subsubsection{Shape Term}
\label{subsubsec::shape}

Registering the point cloud in neighboring frames could fail in long term tracking due to drifting or shifting in point cloud distribution. We remedy this by adding regularization between the shapes of objects and current frame point clouds. 
Specifically, we penalize the inconsistency between the current frame point cloud $\widehat{O_k}$ and the latest shape $M_{k-1}$ by registering them.
In the implementation, we first use the first frame point cloud as initialization, since the initial bounding box has high quality and the related point clouds are accurate.
Then we aggregate new LiDAR points along the process of tracking, and use this updated shape to compute the loss function.
Formally, the Shape Term is computed as Eq.~\ref{eq::shape}. 
$\mathop{\mathcal{A}}_{S}^{M_{k-1}, \widehat{O}_k}$ is the set of associate point pairs, explained in Sec.~\ref{subsubsec::lidar_pa}.

\begin{equation}
    \begin{footnotesize}
        \begin{aligned}
    \mathcal{L}_{Shape} = \frac{1}{|\mathop{\mathcal{A}}_{S}^{M_{k-1}, \widehat{O}_k}|} \sum_{(p, q) \in \mathop{\mathcal{A}}_{S}^{M_{k-1}, \widehat{O}_k}} ||(S_k-\widetilde{S}_1)\otimes (p, \widetilde{S}_1) - q||_2^2
        \end{aligned}
    \end{footnotesize}
    \label{eq::shape}
\end{equation}  

\subsubsection{Motion Consistency Term}
\label{subsubsec::motion_consistency}

The objects cannot be treated as point mass, so their movements have to follow certain constraints.
For simplicity, we enforce the most general constraint: their directions should be consistent with their own headings.   
The corresponding term is computed as Eq.~\ref{eq::motion_consistency_term}, 
where $v = [\Delta x_{k}, \Delta y_{k}]$ represents the velocity.
\begin{equation}
    \label{eq::motion_consistency_term}
    \begin{small}
    \begin{aligned}
    \mathcal{L}_{MC}=\left\|||v||_2\cos(\frac{\theta_{k-1} + \theta_{k}}{2})-\Delta x_{k}\right\|_2^2
\end{aligned}
    \end{small}
\end{equation}

\subsubsection{Motion Prior Term}
\label{subsubsec::motion_prior}

To combine the motion model into the optimization framework, we design the ``Motion Prior'' term. 
It penalizes the inconsistency between the optimization results $\Delta S_k$ and the prediction from motion models $\Delta S_{prior}$.
This term encourages the consensus between the latest optimization and trajectory histories.
The computation of Motion Prior Term is as Eq.~\ref{eq::motion_prior_term}.
\begin{equation}
    \label{eq::motion_prior_term}
    \begin{small}
        \begin{aligned}
	\mathcal{L}_{MP}=\left\| \Delta S -\Delta S_{prior} \right\|_2^2
        \end{aligned}
    \end{small}
\end{equation}
 
\vspace{-3px}
\subsection{Design Choices}
\vspace{-3px}
\label{subsec::design}

We identify several key implementation factors that are helpful for the performance of state estimation.

\subsubsection{Ground Removal}
The ground points are irrelevant with the vehicle's states and shapes, and they usually confuse the algorithm.
So we remove the ground points using~\cite{zermas2017fast} as a pre-processing operation.

\subsubsection{Subshape}
Due to the sparsity of LiDAR point clouds, it is challenging to only use single frame for registration.
In the ICP term, we can enhance the performance by overlaying the LiDAR points in nearby frames.
In the implementation, we overlay additional 2 frames $O_{k-3:k-2}$ onto $\widehat{O}_{k-1}$ in Eq.~\ref{eq::icp}.

\subsubsection{LiDAR Point Association}
\label{subsubsec::lidar_pa}
As a prerequisite of computing the ICP Term and Shape Term in Sec.~\ref{subsec::factor}, 
the LiDAR points have to be associated according to the rules of nearest neighbors.
As for the ICP term described in Sec.~\ref{subsubsec::icp}, 
we follow the common approach and construct the associated point set $\mathop{\mathcal{A}}_{I}^{O_{k - 1}, \widehat{O}_k}$ as Eq.~\ref{eq::forward}.
In the equation, the function $\mathtt{NN}(p, \mathcal{B})$ means finding the nearest neighbor for point $p$ in the point set $\mathcal{B}$.
\begin{equation}
    \label{eq::forward}
    \begin{footnotesize}
    \begin{aligned}
        \left\{ (x, y) \mid y = \mathtt{NN}(\Delta S_{k}\otimes (x, S_{k-1}), \widehat{O}_{k}), \forall x\in O_{k-1} \right\}
    \end{aligned}
    \end{footnotesize}
\end{equation}

However, as for the Shape Term in Sec.~\ref{subsubsec::shape}, the point clouds $M_{k-1}$ and $\widehat{O}_k$ may differ significantly during the process of tracking.
The outliers of association lead to deviated solutions and degrade the state estimation.
Therefore, we use RANSAC for Shape Term to exclude the outlier pairs.
The result $\mathcal{A}_{S}^{M', \widehat{O}_k}$ is computed as Eq.~\ref{eq::shape_outlier}.

\begin{equation}
    \label{eq::shape_outlier}
    \begin{footnotesize}
    \begin{aligned}
        \mathtt{RANSAC}(\left\{ (x, y) \mid y = \mathtt{NN}((S_{k}-\widetilde{S}_{1})\otimes (x, \widetilde{S}_{1}), \widehat{O}_{k}), \forall x\in M_{k-1} \right\})
    \end{aligned}
    \end{footnotesize}
\end{equation}

\vspace{-12px}
\subsection{Limitations and Future Work}
\vspace{-3px}
For simplicity, our \benchmark\ is currently limited to the rigid type of objects: vehicles.
Extending it to more general types of objects, such as cyclists and pedestrians, 
requires the modeling for the periodically and continuously changing shapes. Using multiple templates or implicit functions~\cite{park2019deepsdf} for their shapes could be the solution.

Combining SOT methods with the multiple detections in different frames also poses great challenges, 
as the quality of detection bounding boxes may vary across different frames.
Therefore, we have to model the quality of them and combine our optimization-based algorithm into the MOT framework. 
Also, the disappearance or occlusion of an object need to be considered in the setting. 
We leave these two issues, and will try to solve them in future work.
\vspace{-3px}
\section{Results and Analyses
}
\vspace{-3px}
\label{sec::results}
\begin{table*}[!ht]
    \centering
    \resizebox{1.0\linewidth}{!}{
    \begin{tabular}{@{\hspace{2.0mm}}l|@{\hspace{2.0mm}}l|@{\hspace{2.0mm}}c@{\hspace{2.0mm}}c@{\hspace{2.0mm}}c|@{\hspace{2.0mm}}c@{\hspace{2.0mm}}c@{\hspace{2.0mm}}c|
        c@{\hspace{2.0mm}}c@{\hspace{2.0mm}}c|@{\hspace{2.0mm}}c@{\hspace{2.0mm}}c@{\hspace{2.0mm}}c@{\hspace{2.0mm}}c}
        \hline
        \multirow{2}{*}{Type} & \multirow{2}{*}{Method} & \multicolumn{3}{c}{All} & \multicolumn{3}{c}{Easy} & \multicolumn{3}{c}{Medium} & \multicolumn{3}{c}{Hard} \\
        \cline{3-14} & & Acc$\uparrow$ & Rob$\uparrow$ & Shape$\downarrow$  &Acc$\uparrow$ & Rob$\uparrow$ & Shape$\downarrow$  & Acc$\uparrow$ & Rob$\uparrow$ & Shape$\downarrow$  & Acc$\uparrow$ & Rob$\uparrow$ & Shape$\downarrow$  \\
        \hline\hline
        \multirow{7}{*}{\shortstack{Model-\\free}} 
        & ICP+MP           & 0.3927 & 0.3734 & 0.1836 & 0.5156 & 0.4997 & 0.1067 & 0.3578 & 0.3340 & 0.1792 & 0.2828 & 0.2638 & 0.2652 \\
        & ICP+MP+S(1)      & 0.5931 & 0.5113 & 0.1226 & 0.7386 & 0.6801 & 0.0612 & 0.5380 & 0.4522 & 0.1218 & 0.4779 & 0.3721 & 0.1851 \\
        & ICP+MP+S(All)    & 0.6027 & 0.5348 & 0.1239 & 0.7479 & 0.6992 & 0.0591 & 0.5613 & 0.4819 & 0.1261 & 0.4789 & 0.3946 & 0.1867 \\
        & \textbf{ICP+MP+S(All)+MC}  & \textbf{0.6146} & \textbf{0.5467} & \textbf{0.1164} & \textbf{0.7496} & \textbf{0.7002} & \textbf{0.0589} & \textbf{0.5743} & \textbf{0.4910} & \textbf{0.1173} & \textbf{0.4959} & \textbf{0.4224} & \textbf{0.1727} \\
        \cline{2-14}
        & w/o GM        & 0.5715 & 0.4994 & 0.1284 & 0.6924 & 0.6450 & 0.0676 & 0.5320 & 0.4512 & 0.1279 & 0.4662 & 0.3783 & 0.1889 \\
        & w/o RANSAC    & 0.5707 & 0.5130 & 0.1276 & 0.7016 & 0.6615 & 0.0705 & 0.5185 & 0.4480 & 0.1239 & 0.4728 & 0.4049 & 0.1883 \\
        & w/o SUB       & 0.6025 & 0.5367 & 0.1221 & 0.7436 & 0.6937 & 0.0595 & 0.5574 & 0.4813 & 0.1199 & 0.4814 & 0.4079 & 0.1871 \\  
        \hline
        \multirow{2}{*}{\shortstack{Model-\\based}} 
        & Det & 0.7574 & 0.6184 & 0.0714 & 0.7850 & 0.6918 & 0.0589 & 0.7409 & 0.5993 & 0.0780 & 0.7423 & 0.5508 & 0.0774 \\   
        & Det+SOT  & 0.7690 & 0.6329 & 0.0644 & 0.8017 & 0.7092 & 0.0544 & 0.7597 & 0.6169 & 0.0637 & 0.7398 & 0.5587 & 0.0753 \\   
        \hline
    \end{tabular}
    }
    \caption{Ablation studies. $\uparrow$($\downarrow$) means the performance is better with larger(smaller) values. 
    Optimization terms: \textbf{ICP} -- ICP Term; \textbf{S} -- Shape Term, \textbf{S(1)} -- only using the first frame point cloud, \textbf{S(all)} -- using the updated shape; 
    \textbf{MC} -- Motion Consistency Term, \textbf{MP} -- Motion Prior Term. 
    Implementation details:
    \textbf{GM} -- ground removal, \textbf{RANSAC} -- using RANSAC, \textbf{SUB} -- subshape.
    \textbf{Det} and \textbf{Det+SOT} are covered in Sec.~\ref{subsec::model_based}.}
    \vspace{-15px}
    \label{tab::ablation}
\end{table*}

\begin{figure}
	\centering
	\includegraphics[width=0.90\columnwidth]{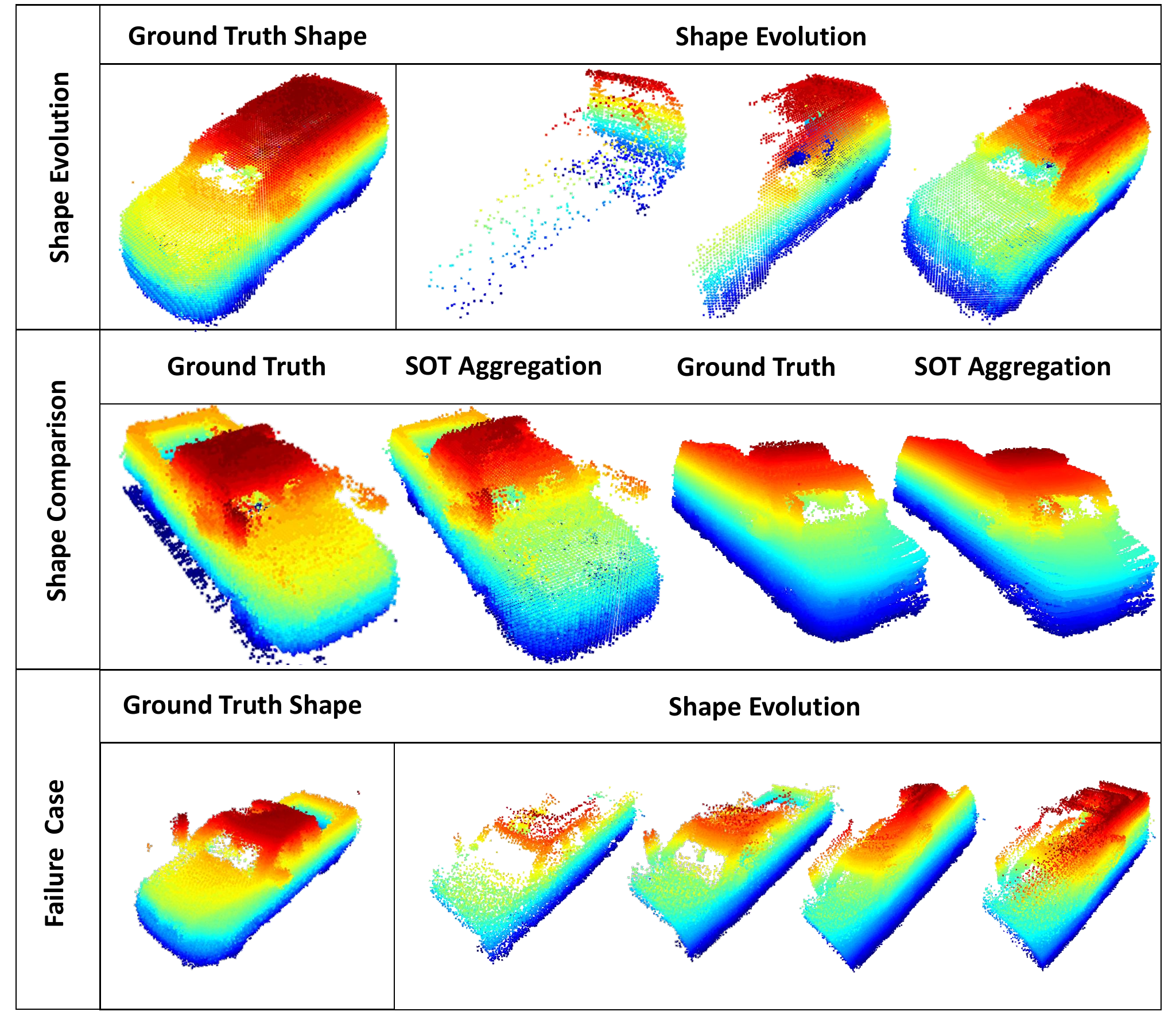}
	\caption{Visualization SOT results.
		\textbf{1}. The evolution of shapes during the tracking process.
		\textbf{2}. Comparison between the results from SOT and ground truth.  
		\textbf{3}. Visualization of failure cases.}
	\label{fig::visualizations}
	\vspace{-15px}
\end{figure}

\subsection{Quantitative Results}
\label{subsec::quantitative}
\vspace{-3px}

Table~\ref{tab::ablation} evaluates the contributions of each individual factor. We start from the baseline of using ICP and motion model, then add the other terms one by one.
When ablating the optimization factors from Sec.~\ref{subsec::factor}, we remove them separately from the strongest \tracker\ to show their benefits.  

Compared to the baseline of using ICP and motion model only (ICP+MP), adding the Shape-Term (S(1), using the first frame point cloud as shape in the Shape Term) significantly improves the performance.
During this process, aggregating shapes along the process of tracking (S(All), using aggregated shape in the Shape Term) assist the state estimation even further.
On top of the point cloud related factors, the regularization on motion consistency (MC) is helpful especially on medium and hard cases.
As for the design choices in Sec.~\ref{subsec::design}, removing any of them greatly affects both the accuracy and robustness, 
where Ground Removal and RANSAC are the most critical ones.

\vspace{-3px}
\subsection{Comparison with Model-based Methods}
\vspace{-3px}
\label{subsec::model_based}

\subsubsection{Compare with Model-based Baseline}

To demonstrate the effectiveness of our model-free approach, we compare our method with one common model-based baseline: utilizing Kalman filters to associate the detections, and then directly output the associated bounding boxes. Specifically, we use \textit{Point Pillars}~\cite{pointpillars} for detection. Then we change the detection bounding boxes' sizes to ground-truth following LiDARSim~\cite{LiDARsim} for fair comparison with SOT methods, since other SOT methods have ground-truth size in the first frame.

Contrasting Row 4 and ``Det'' in Tab.~\ref{tab::ablation}, we observe that our \tracker~ can approach the performance of ``Det'' on the easy set without external training data, even slightly better on robustness, while has large disadvantages on the medium and hard sets. The results are reasonable because model-based methods could benefit from the large-scale training set, while model-free methods simply don't use it, but should be more robust on the rare and unseen cases.

\subsubsection{Combine Model-free and Model-based Methods}

Furthermore, we try to incorporate the detections into our framework by \textit{Detection Term}. Briefly speaking, it selects the detection with the largest IOU with the motion model prediction, while enforcing a minimum score of 0.5 and IOU of 0.1. If no bounding box fulfills the requirement, this term is discarded. The term is illustrated in Eq.~\ref{eq::det_term}, where $S_{\mathtt{D}}$ is the state for selected detection. This hybrid method is the ``Det+SOT'' in Tab.~\ref{tab::ablation}. Compared to the ``Det'' row, ``Det+SOT'' consistently outperforms, proving our point.
\begin{equation}
    \label{eq::det_term}
    \begin{small}
        \begin{aligned}
\mathcal{L}_{Det} = \left\| \Delta S_k - (S_{\mathtt{D}}-S_{k-1}) \right\|^{2}_{2}
        \end{aligned}
    \end{small}
\end{equation}

\vspace{-3px}
\subsection{Analyses}
\vspace{-3px}

\subsubsection{\textbf{Vehicle Shape Visualization}}

Following the approach in~\cite{Held-Combining,IROS-Continuous-time}, we visualize the aggregated point clouds in Fig.~\ref{fig::visualizations} to indicate the quality of tracking. 
Moreover, the vehicle banks constructed during this process are also beneficial to a series of tasks discussed in Sec.~\ref{sec::downstream}. 
In Fig.~\ref{fig::visualizations} and Fig.~\ref{fig::cover}, 
the evolution of point clouds and the quality of shapes prove that our algorithm can produce high quality state information and 3D shapes.
This further poses the necessity and effectiveness of using LiDAR sequences for creating 3D shapes.

\subsubsection{\textbf{Shape Completion $\not=$ State Estimation}}
\label{subsubsec::shape_not_track}

Although in Table.~\ref{tab::ablation}, better shapes generally come along with better state estimation performance,
we argue that these two are not equivalent tasks.
The LiDAR points are aggregated until the last frame in the experiments of Table~\ref{tab::ablation},
but this is not the optimal approach and leads to worse shapes on the failures of state estimation, just as the last row in Fig.~\ref{fig::visualizations}.
Therefore, ideal shape completion needs to reason whether and how to aggregate the point clouds, thus differs from mere state estimation.
We will investigate this direction in future study.
\vspace{-3px}
\subsection{Challenging Scenarios}
\vspace{-3px}
\subsubsection{Point Cloud Sparsity} 
In Table~\ref{tab::ablation}, the performance gaps among the three difficulty levels reflect the challenge of sparse point clouds.
The hard set has less LiDAR points, which causes inaccurate registration at the beginning and leads to lower performance.
In long-term tracking, the ubiquitous cases of occlusion or distant objects indicate the needs for handling this challenge. 

\subsubsection{Drifting}
Accumulated error is well known in SLAM and commonly exists in odometry based approaches. 
The errors can accumulate and cause drifting over time if without absolute observation or extra constraints like loop closure. 
We alleviate this by adding the shape terms, which treats the shape as ``Map'' and enforce its consistency with the observed point cloud. 
Although the terms are effective in many cases as in Table~\ref{tab::ablation}, 
drifting still exists as in Fig.~\ref{fig::cover} and requires future research on it.

\subsubsection{Abrupt Motion Change}
Accurate motion initialization is crucial since it provides the initial point set correspondence.
Although current motion models are able to adapt to most cases,
they still struggle where the objects change the motions greatly in short periods,
as the incorrectly associated point clouds can quickly lead to failures.
We will explore more advanced motion prediction methods, and close the loop of state estimation and motion prediction in the future.

\section{Experiments on Downstream Applications}
\label{sec::downstream}
\vspace{-3px}

\subsection{Motion Ground Truth Generation}
\label{subsec::motion_generation}
\vspace{-3px}

\begin{figure}
    \centering
    \includegraphics[width=0.90\columnwidth]{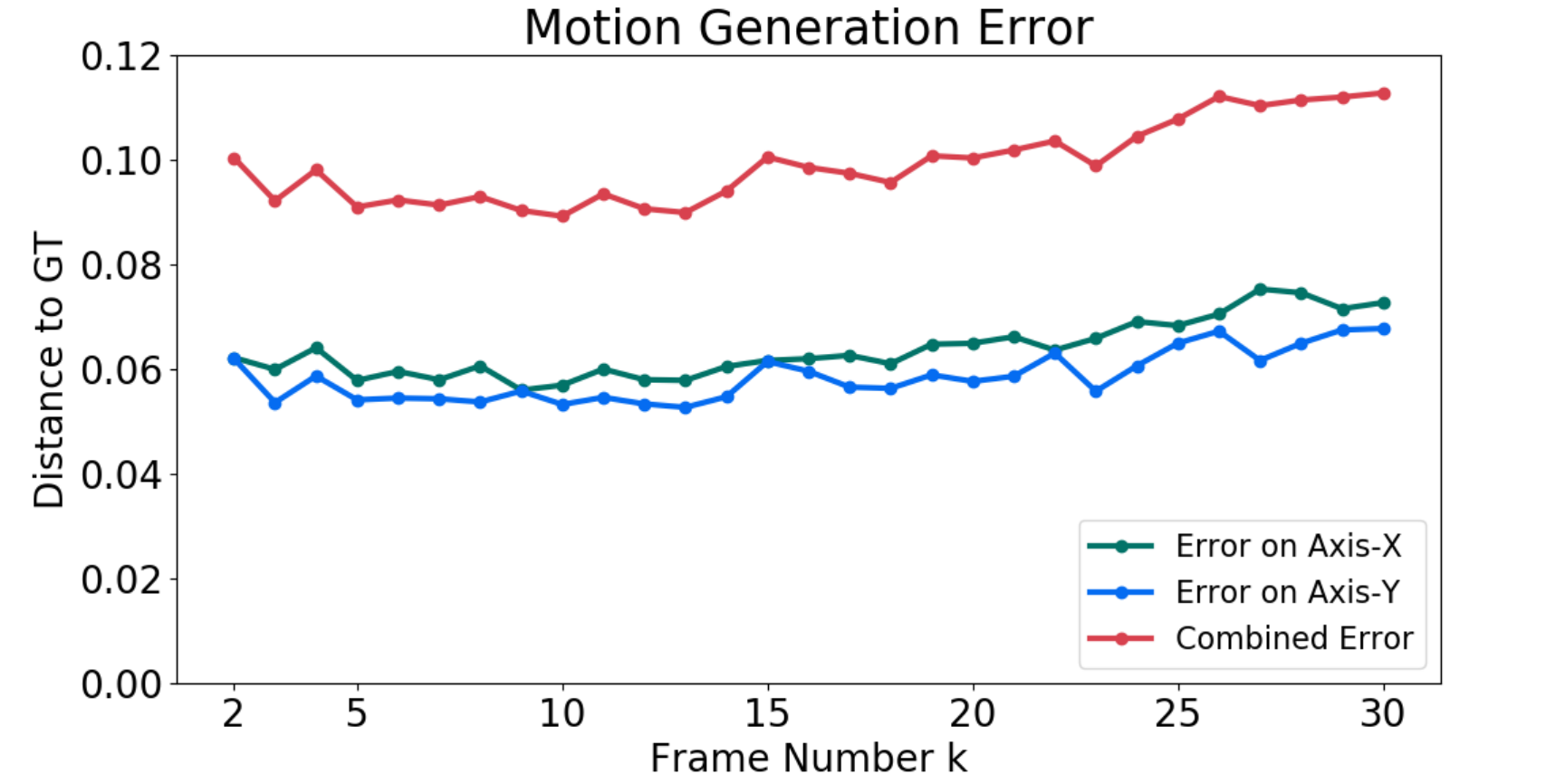}
    \caption{Motion generation results.
    The average distance (meters) between the motion from ground-truth and our algorithm at frame k.
    }
    \label{fig::motion_generation}
    \vspace{-15px}
\end{figure}

\tracker\ can estimate the motion of a designated vehicle in wild point cloud sequences.
Therefore, it is capable of providing motion data for prediction and planning systems at a low cost.
As in Fig.~\ref{fig::motion_generation}, the deviation of generated motion is less than 10cm within 30 frames, which is even smaller than the size of a finger.
With Argoverse~\cite{argoverse} providing 30 data points for the motion prediction task, we believe that our system meets the standard.

\vspace{-3px}
\subsection{Optical Flow Annotation}
\vspace{-3px}

With the estimated shapes $M$ and states $S_{k:k+1}$, we can annotate optical flows for the tracked objects in corresponding images $I_{k:k+1}$.
For each point $m\in M$, we compute its location $(qx_{k}, qy_{k})$ on $I_{k}$ by moving the shape according to $S_k$, then project it onto the image plane.
In the same way, we compute $(qx_{k+1}, qy_{k+1})$ on image $I_{k+1}$, then the optical flow for pixel $(qx_{k}, qy_{k})$ is $(qx_{k+1}-qx_k, qy_{k+1}-qy_k)$. 
In Fig.~\ref{fig::cover}, our annotated optical flow can act as high quality pseudo-ground-truth.
Such an application is infeasible for the single frame point cloud (the left image), as it contains neither enough points nor the motion information for annotating the optical flow.
In KITTI~\cite{KITTI-Flow}, researchers require all three of LiDAR data, CAD models and ground-truth 3D bounding boxes to infer the optical flows of vehicles,
while our algorithm only requires an initial bounding box and raw LiDAR scenes.
Therefore, our method can reduce the cost of optical flow annotation and scale it up.

\vspace{-3px}
\subsection{Simulating LiDAR Scans Using Vehicle Bank}
\vspace{-3px}

\begin{figure}
	\centering
	\includegraphics[width=0.90\columnwidth]{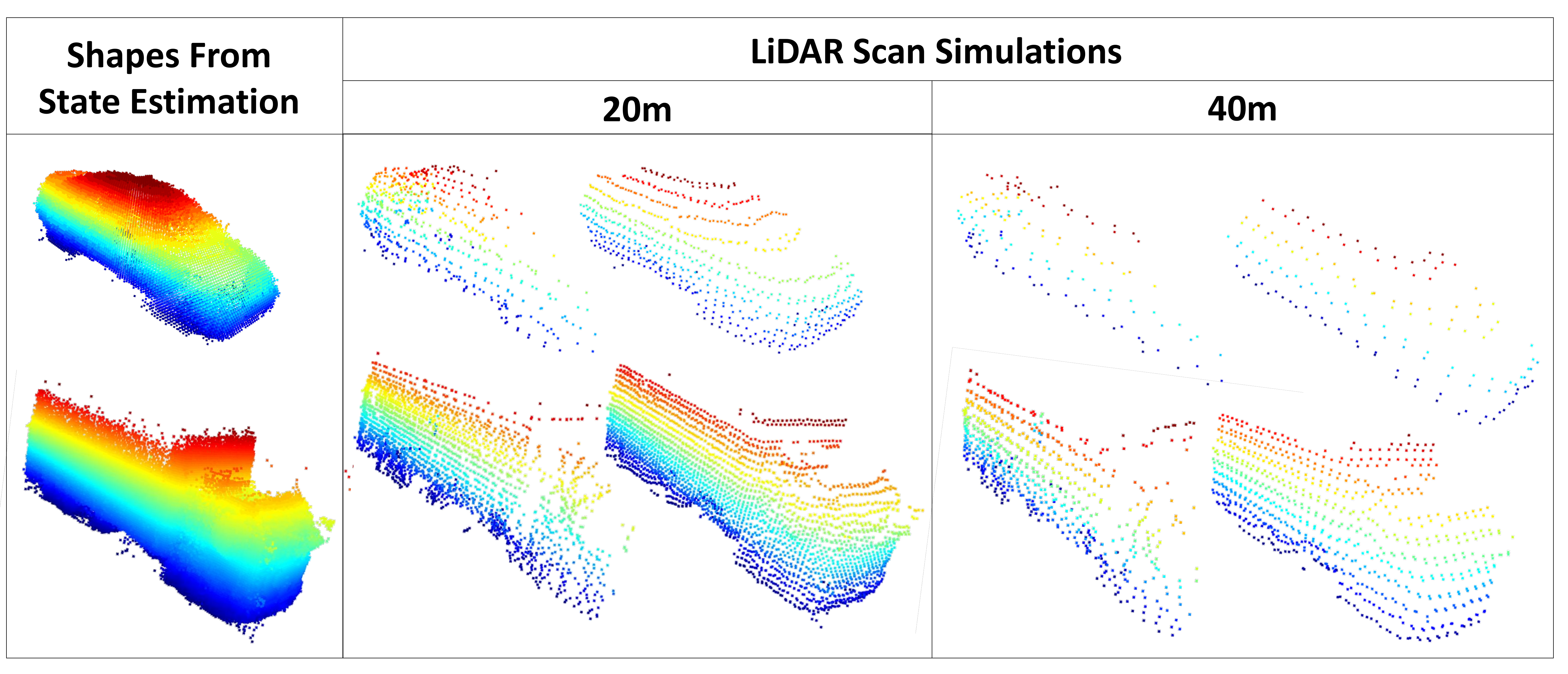}
    \caption{
    LiDAR scans created using our completed shapes and physical simulation. 
    The visualization includes two types of vehicles: car and truck.
    The LiDAR simulation fits different angles and different distances (20m and 40m).
    }
    \vspace{-16px}
	\label{fig::lidar_sim}
\end{figure}

In LiDARsim~\cite{LiDARsim} researchers propose to simulate single LiDAR scans for autonomous driving by creating object shapes with manually labeled 3D locations.
As mentioned in LiDARsim~\cite{LiDARsim}, augmenting the training data with the simulation can improve the performance of perception systems.
In Fig.~\ref{fig::lidar_sim} we visualize the results of our LiDAR scan simulation on different types of vehicles, angles, and distances.
For both types of vehicles, we can build up high quality shapes and simulated scans, which proves the effectiveness of our model-free SOT approach. 
Moreover, unlike LiDARsim~\cite{LiDARsim}, which requires exhausted annotation of objects, we only require an initial bounding box and raw point clouds.
Thus, we make the process cheaper and more scalable, and produce unlimited LiDAR data with minimum human annotation.



\vspace{-3px}
\section{Conclusion}
\vspace{-3px}

In this paper, we calls for the attention on a new setting of object tracking in point cloud sequences: 
model-free single-object tracking.
This new setting targets on the drawbacks of model-based detection and MOT methods, and can benefit a series of applications.
To facilitate the research, we construct the \benchmark\ from WOD and propose a strong algorithm called \tracker.
According to the qualitative and quantitative results, 
our method not only achieves strong baseline on state estimation,
but also generalizes its estimated states and 3D shapes to the applications of simulating LiDAR scans, optical flow annotation, and motion data generation.
\vspace{-5px}



{\small
\bibliographystyle{ieee}
\bibliography{reference}
}

\end{document}